\documentclass[letterpaper, 10 pt, conference]{ieeeconf} 	%
\IEEEoverridecommandlockouts	%
\usepackage{graphics}		%
\usepackage{epsfig}			%
\usepackage{graphicx}
\usepackage{amsmath}		%
\usepackage{amssymb}		%
\usepackage{hyperref}
\usepackage{algorithm}		%
\usepackage{algpseudocode}	%
\usepackage[table]{xcolor}
\usepackage{array}
\usepackage{float}
\usepackage{color}
\usepackage{booktabs}
\usepackage{amssymb}
\usepackage{bm,amsmath}
\usepackage{subcaption}
\graphicspath{ {fig/} }
\usepackage{hyperref}
\usepackage{xr}
\graphicspath{ {Images/} }

\usepackage{textcomp}

\usepackage{siunitx}

\usepackage{epstopdf}
\usepackage{color}

\newcolumntype{C}[1]{>{\centering\let\newline\\\arraybackslash\hspace{0pt}}m{#1}}

\usepackage{subcaption}

\usepackage{microtype}
\usepackage{amsmath,mathtools} %

\DeclareMathOperator*{\argmin}{\arg\!\min}

\title{\LARGE \bf Whole-Body Geometric Retargeting for Humanoid Robots}
\author{Kourosh Darvish$^{1}$$^{*}$, Yeshasvi Tirupachuri$^{1}$$^{2}$$^{*}$, Giulio Romualdi$^{1}$$^{2}$, \\ Lorenzo Rapetti$^{1}$$^{3}$, Diego Ferigo$^{1}$$^{3}$, Francisco Javier Andrade Chavez$^{1}$, Daniele Pucci$^{1}$
\thanks{$^{*}$ Equal author contribution. This work is supported by \href{http://itn-pace.eu/}{PACE} project, Marie Skłodowska-Curie grant agreement No. 642961 and \href{https://andy-project.eu/}{An.Dy} project which has received funding from the European Union\textquotesingle s Horizon 2020 Research and Innovation Programme under grant agreement No. 731540.}
\thanks{$^{1}$ Dynamic Interaction Control, Istituto Italiano di Tecnologia, Genova, Italy {\tt\small name.surname@iit.it}}
\thanks{$^{2}$ DIBRIS, University of Genova, Genova, Italy}
\thanks{$^{3}$ School of Computer Science, The University of Manchester, Manchester, United Kingdom}
}

\begin{document}

\maketitle
\thispagestyle{empty}
\pagestyle{empty}

\begin{abstract}

Humanoid robot teleoperation allows humans to integrate their cognitive capabilities with the apparatus to perform tasks that need high strength, manoeuvrability and dexterity. This paper presents a framework for teleoperation of humanoid robots using a novel approach for motion retargeting through inverse kinematics over the robot model. The proposed method enhances scalability for retargeting, i.e., it allows teleoperating different robots by different human users with minimal changes to the proposed system. Our framework enables an intuitive and natural interaction between the human operator and the humanoid robot at the configuration space level. We validate our approach by demonstrating whole-body retargeting with multiple robot models. Furthermore, we present experimental validation through teleoperation experiments using two state-of-the-art whole-body controllers for humanoid robots.

\end{abstract}

\section{Introduction}
\label{sec:Introduction}

Teleoperation stands for operating from distance, in which it extends the human capability to operate a robot remotely, where the human is unable to reach owing to the time and space constraints or the dangers posed by hazardous environments \cite{Hokayem2006}. Moreover, the perception and decision making capabilities of current robotic systems are still limited preventing them from acting autonomously out of laboratory settings and in real-world conditions. Teleoperation plays an important role in a wide range of applications including  manipulation in hazardous environments \cite{Shimoga1993}, \cite{Trevelyan2016}, telepresence \cite{Tachi1989}, telesurgery \cite{Burgner2015}, and space exploration \cite{Pedersen2003}. Teleoperation is considered as a type of human-robot interaction at distance, where for an effective and efficient mission, a bilateral communication is paramount \cite{Goodrich2007}. Given this perspective, the human and the robot establish a team, in which the goal of the teleoperated robot is the same as the human operator. Furthermore, teleoperation brings together the excellent cognitive capabilities of humans and the physical strength of the robotic system together \cite{Zucker2015}.

\begin{figure}[ht]
	\centering
	\includegraphics[scale=0.1]{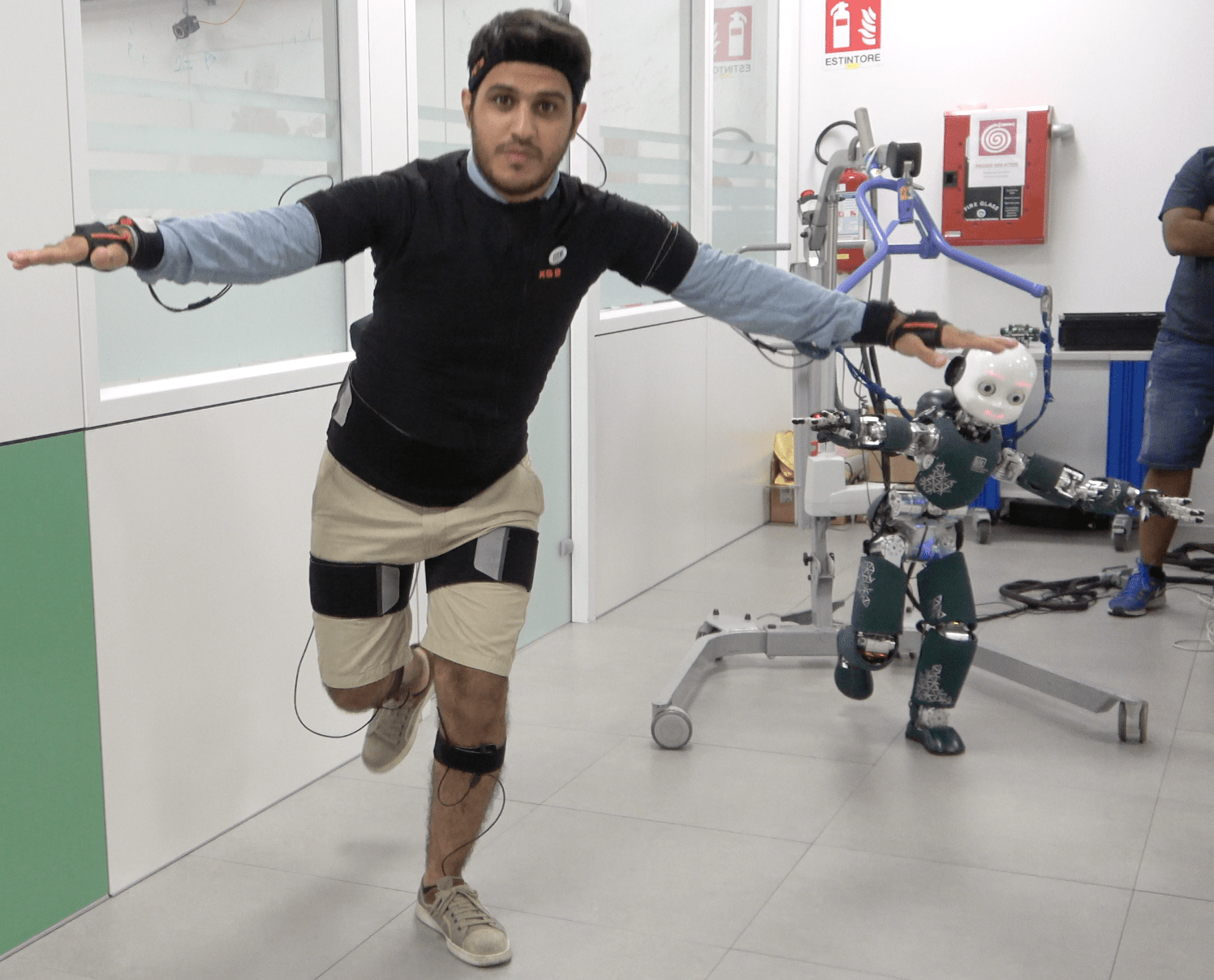}
	\caption{Whole-body retargeting example scenario}
	\label{fig:whole-body-retargeting-example}
\end{figure}

Humanoid robots are designed based on the idea of anthropomorphism and unlike serial manipulators, they have higher manoeuvrability and manipulation capabilities \cite{Ishiguro2018}. Hence, they facilitate higher capabilities during teleoperation. At the same time, the complexity of humanoid robots offers more challenges for teleoperation particularly in unstructured dynamic environments designed for humans. The level of autonomy, team organization and, the information exchange between the operator and the robot are some of the vital aspects in teleoperation performance to ensure successful task completion \cite{Beer2014, Steinfeld2006}. The level of autonomy ranges from being a semi-autonomous robot at the symbolic or the action level (high-level teleoperation) \cite{Goodrich2007, Hokayem2006} to complete control of the robot at the kinematic and the dynamic level (low-level teleoperation), either in the robot's configuration space or task space.
A core component of the low-level teleoperation system is the human motion retargeting to a robot. An example scenario of whole-body retargeting of human motion to a humanoid robot is shown in Fig.~\ref{fig:whole-body-retargeting-example} where each limb of the robot mimics the motion of the human limbs.

Two of the most studied teleoperation paradigms in literature are: 1) master-slave; and 2) bilateral systems. Under master-slave teleoperation paradigm, the flow of information is unidirectional from the human to the robot, while under bilateral teleoperation paradigm there is an exchange of information between the human and the robot. In particular, haptic feedback to the human from the robot \cite{Ishiguro2017, Wang2015}. Teleoperation systems that involve humans in the control loop at the kinematic and dynamic level should have the prime objectives of situational awareness and transparency, i.e., the human operator experiencing the remote environment of the teleoperated robot as holistically as possible, while maintaining the stability of the closed-loop system \cite{Lichiardopol2007, Hokayem2006}. Delays and information loss are some of the crucial problems with this approach that affect the transparency and stability of teleoperation greatly \cite{Hokayem2006, Lichiardopol2007}.  Different approaches such as Lyapunov stability analysis \cite{Islam2015, Chopra2003}, passivity based control \cite{Chopra2003} have been employed to address these limitations. However, these methods are studied extensively with manipulators and the stability measures for humanoid robot teleoperation are not well established \cite{Ramos2018TRO}.

The research on teleoperation of humanoid robots can be broadly classified into three categories: upper body teleoperation, lower body teleoperation, and whole-body teleoperation. In upper body teleoperation, the mapping of the human motion to the robot motion is considered at the kinematic level. Inverse Kinematic (IK) and nonlinear optimization approaches are the common methods for teleoperation scenarios. Using inverse kinematics methods, human joint angles or velocities are computed and mapped at configuration space to the corresponding joints of the humanoid robot, taking into account the robot limitations \cite{Liarokapis2013, Ayusawa2017, Stanton2012}. Nonlinear optimization methods map the human's hand motions at task space to the desired trajectory of the humanoid robot end-effector motions \cite{Elobaid2018, Liarokapis2013}. Alternatively, a data-driver mapping between the human and the robot arm is proposed in \cite{Pierce2012}. In these cases, some consider the effect of the change of center of mass (COM) in robot lower body motion and the robot's balance \cite{Elobaid2018}, while others do not and therefore, the risk of robot falling down increases. So, concerning the lower body teleoperation of humanoid robots, the aspects of stability and locomotion have higher precedence over retargeting of all the lower limbs. A more detailed description of such methods are discussed in \cite{Romualdi2018, Feng2015}.

Coming to the topic of whole-body teleoperation of humanoid robots, the key challenge is to control the robot such that it does not fall while keeping its manoeuvrability and manipulability high, so that the human and the robot team can successfully perform a given task. The balance of the robot is achieved by either keeping the Center of Mass (CoM) inside the support polygon or maintaining the net momentum about the Center of Pressure (CoP) to zero \cite{Penco2018, Ishiguro2018}. Although, a set of safety limitations are considered to maintain the robot's stability, multi-link dynamic contacts are not considered in \cite{Ishiguro2018}. Therefore, they can not handle tasks which needs force exchange with the environment or compensate for external disturbances. An attempt to solve this problem is presented in \cite{Ramos2018TRO} with simulations which uses the natural frequencies of human and robot models in the feedback law, and synchronize their motions to compute the robot balancing and stepping strategies. Differently from the described methods, a data-driven approach for whole-body retargeting in a physics based animation environment is proposed by the authors of \cite{Peng2018}.

One of the obvious shortcomings of the teleoperation systems proposed in literature is the lack of ability to easily adapt the system for different human users and humanoid robots with different geometries, kinematics, and dynamics. The possibility to perform human motion retargeting without considering major design changes becomes limited. The system designer often has to spend time and effort in finding a new model of the human to be used during motion retargeting step such as IK based approaches, therefore the \textit{usability} and \textit{scalability} of the proposed teleoperation system decreases.

This paper presents a novel framework for whole-body retargeting and teleoperation of a humanoid robot that enhances the scalability to multiple human operators or multiple robot models. Our approach provides anthropomorphic references for humanoid robot joints in real-time based on the human limb motion measures, independently from the human body dimensions, by directly using the robot model. The proposed approach is validated by extensive whole-body retargeting and teleoperation experiments.

The rest of the paper is organized as follows: Section \ref{sec:background} introduces the basic notations, robot modelling, and an overview of motion retargeting. Section \ref{sec:methods} presents our whole-body retargeting architecture. Section \ref{sec:retargeting-experiments-results} describes the whole-body retargeting experiments and highlights the results validating our approach. Section \ref{sec:teleoperation-experiments-results} shows the experiments and results of whole-body teleoperation with two state-of-the-art whole-body controllers for humanoid robots. Section \ref{sec:conclusions} provides the conclusions and hints at our future work. 

\section{Background}
\label{sec:background}

\subsection{Notations \& Modeling}
\label{sec:notation-modeling}

The inertial frame of reference is denoted by $\mathcal{I}$.
Given two frames, $\mathcal{A}$ and $\mathcal{B}$, $\prescript{\mathcal{A}}{}{R}_B \in SO(3)$ represents the rotation matrix between the frames, i.e., given two vectors $\prescript{\mathcal{A}}{}{p}, \prescript{\mathcal{B}}{}{p} \in \mathbb{R}^3$ respectively expressed in $\mathcal{A}$ and $\mathcal{B}$, the rotation matrix $\prescript{\mathcal{A}}{}{R}_\mathcal{B}$ is such that $\prescript{\mathcal{A}}{}{p} = \prescript{\mathcal{A}}{}{R}_\mathcal{B} \prescript{\mathcal{B}}{}{p}$. 
The \textit{skew-symmetric} operation of a matrix is defined as $\text{sk}(A) := (A - A^\top)/2$, and the \textit{vee} operator ${.}^{\vee}$ maps a skew-symetric matrix  from $SO(3) \to \mathbb{R}^3$. Humans and humanoid robots are considered to be multibody floating base systems, i.e., none of the links has an \textit{a priori} constant pose with respect to the inertial frame \cite{Featherstone2007, latella2018towards}. Superscripts $.^{H}$ and $.^{R}$ corresponds to a quantity of the human and the robot respectively.
The configuration of the system can be determined by the triplet $q$ that contains the position $\prescript{\mathcal{I}}{}{p}_\mathcal{B}$ and the orientation $\prescript{\mathcal{I}}{}{R}_\mathcal{B}$ of the base frame and the vector of joint values $s$ that highlight the shape of the robot.
The velocity of the multibody system is represented by the triplet $ \nu$ composed by the linear $\prescript{\mathcal{I}}{}{\dot{p}}_\mathcal{B}$ and angular velocity $\prescript{\mathcal{I}}{}{\omega}_\mathcal{B}$ of the base frame with respect to the inertial frame along with the vector of joint velocities $\dot{s}$.
The Jacobian $J_{\mathcal{A}}$ is the map between the robot velocity and the linear and angular velocities of the frame $\mathcal{A}$, i.e.:

\begin{equation}
\prescript{\mathcal{I}}{}{v}_\mathcal{A} = J_{\mathcal{A}}(q) \nu.
\label{eq:jacobian}
\end{equation}
Jacobian matrix $J_{\mathcal{A}}$ is composed of linear part $J_{\mathcal{A}}^{l}$ and angular part $J_{\mathcal{A}}^a$.
Velocity vector of a frame $v$ are made-up of linear $\dot{p}$  and  angular parts $\omega$.

\subsection{Kinematic Motion Retargeting}
\label{sec:kinematic-retargeting}

The two main methods for the retargeting of human motions at the kinematic level are the configuration space retargeting and the task space retargeting. 

\subsubsection{Configuration space retargeting}

\begin{figure}[t]
	\centering
	\includegraphics[width=\columnwidth]{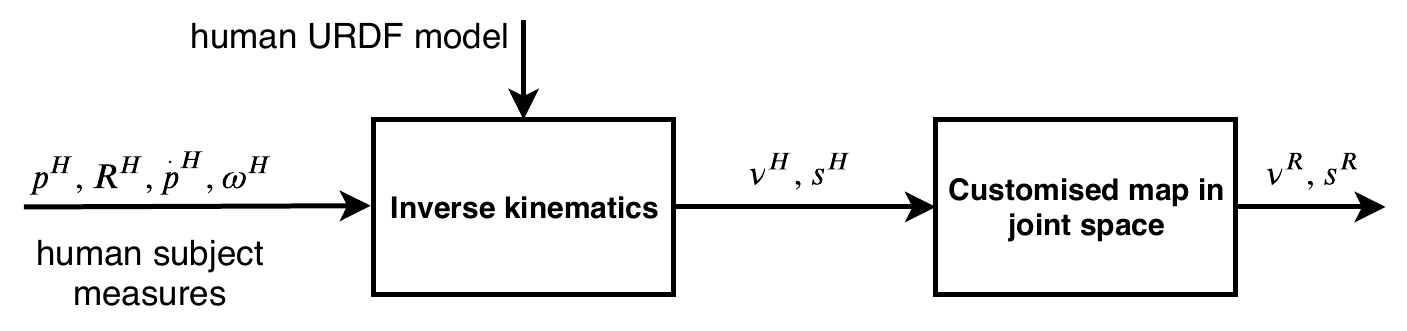}
	\caption{Typical configuration space retargeting scheme.}
	\label{fig:configuration-retargeting}
\end{figure}

The architecture shown in Fig.~\ref{fig:configuration-retargeting} represents a typical configuration space retargeting scheme \cite{Ayusawa2017, Penco2018}. The measurements of the human motion are given as input to an inverse kinematics based method along with the human model. On the output side we retrieve the human joint angles and velocities $s^{H}, \nu^{H}$. Later, a mapping step morphs them to the robot joint angles and velocities $s^R, \nu^R$.

Some of the key limitations of this approach are: i) \textit{finding a customized mapping:} we should apply the constraints of the robot joints to $s^H, \nu^H$, find a customised offset and scaling factor for each of the robot joint with respect to the corresponding human joints; ii) \textit{dissimilarity of the human and the robot kinematics:} the robot kinematics can be different from the humans, for example the human shoulder consists of a spherical joint, while robot's shoulder are three revolute joints with different order. Therefore, at this step we should use forward kinematics to find the relative rotation between the chest link frame and the upper arm link frame of the human, and then apply inverse kinematics to find the robot's joint angles and velocities; iii) \textit{different human kinematics:} different human subjects have different physical properties, which results in different human models.

\subsubsection{Task space retargeting}

\begin{figure}[b]
	\centering
	\includegraphics[width=\columnwidth]{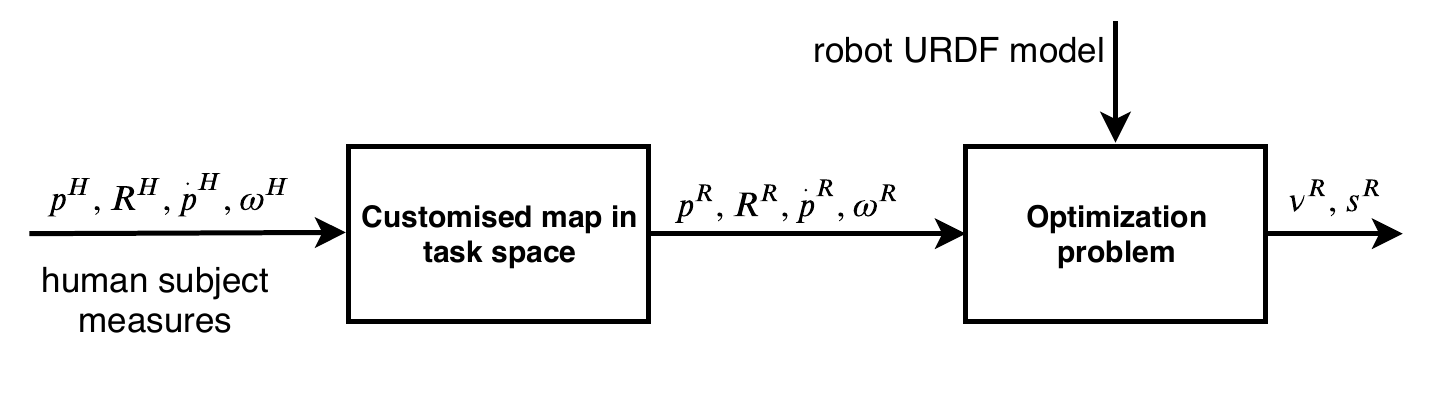}
	\caption{Typical task space retargeting scheme.}
	\label{fig:task-retargeting}
\end{figure}

The architecture shown in Fig.~\ref{fig:task-retargeting} represents a typical task space retargeting scheme. In this approach, the human link measurements in cartesian space are mapped to the robot's cartesian space at the first step \cite{Elobaid2018, Ishiguro2018}. A general attempt for such a mapping is a fixed proportion between the human and robot's geometry, e.g., the human's wrist rotation is mapped equally to robot end effector orientation $^{I}R_{wrist}^R = {}^{I}R_{wrist}^H$ or for the case of robot's end-effector position we have $^{shoulder}p_{wrist}^R= \gamma \ ^{shoulder}p_{wrist}^H$.
A heuristic to find $\gamma$ is provided in \cite{Elobaid2018}, in which it is defined as $\gamma = \frac{\text{robot's arm length}}{\text{human's arm length}}$. Later, the optimization problem, i.e., inverse kinematics, is solved with the robot's model.

Some of the key limitations of this approach are: i) \textit{workspace or precision limits:} the workspace of the robot may be narrowed ($\gamma \leq 1$) for reaching some of the far points or the precision is lost in the case $\gamma \geq 1$ for fine manipulation tasks; ii) \textit{robot's internal configuration dissimilarity:} the robot internal configuration may not be similar to the human, i.e., the degrees-of-freedom problem. It causes psychological discomfort, as the user or people who are interacting with the robot may not predict the robot's motions because of non-anthropomorphic motions that depend on the parameters of the optimization problem \cite{Liarokapis2013}. Moreover, the precise control of internal configuration becomes essential when the robot acts in cluttered environments to avoid obstacles.

\section{Methods}
\label{sec:methods}

\begin{figure*}[t]
	\centering
	\includegraphics[width=\textwidth]{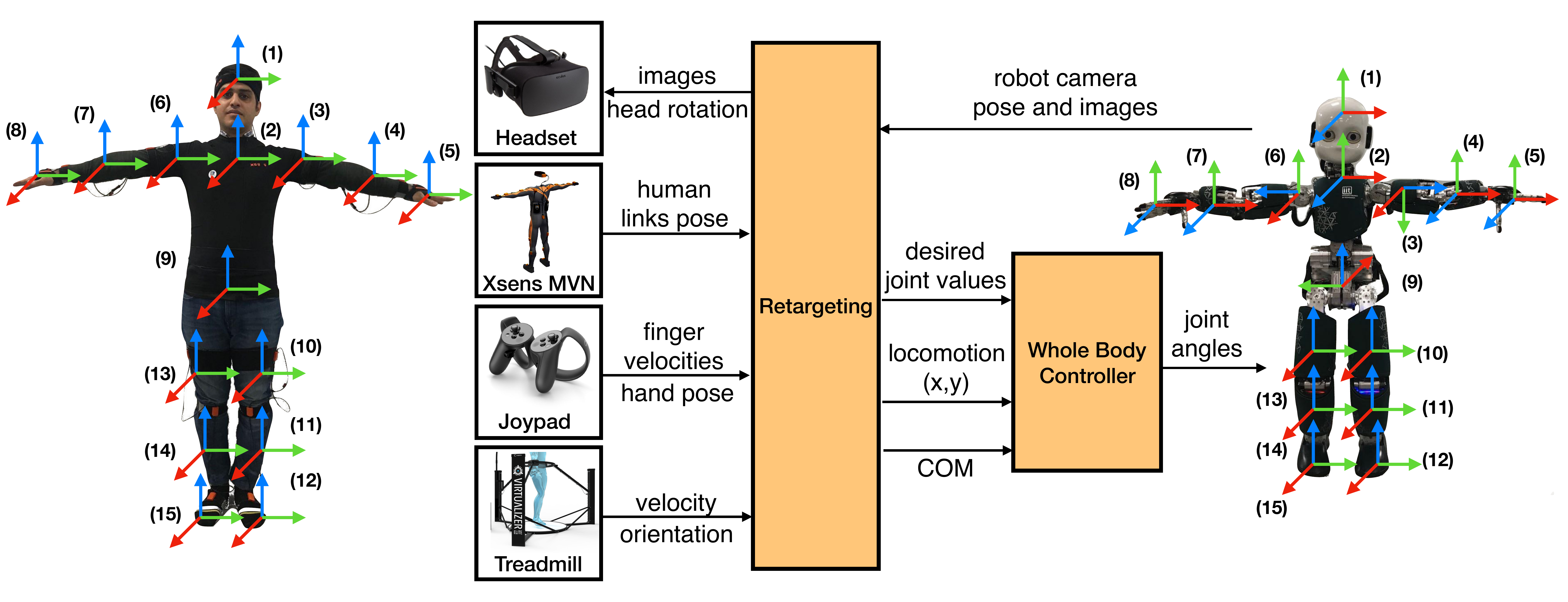}
	\caption{The architecture of the whole-body teleoperation with active human motion retargeting.}
	\label{fig:architecture}
\end{figure*}

\subsection{whole-body Teleoperation Architecture}
\label{sec:RetargetingStructure}

We propose a whole-body teleoperation architecture as shown in Fig. \ref{fig:architecture}.
The human user receives visual feedback from the robot environment by streaming the robot camera images through the \emph{Oculus Headset}.
The robot hands are controlled via the \emph{Joypads}.
The human locomotion information, i.e., the linear and the angular velocities are obtained from the \emph{Cyberith Virtualizer VR Treadmill}. %
Additionally, the human wears a sensorized full body suit from \emph{Xsens technologies} to obtain the kinematic information of various human links with respect to the inertial frame of reference.

\subsection{Kinematic Whole-Body Human Motion Retargeting}
\label{sec:human-motion-retargeting}

\begin{figure}[t]
	\centering
	\includegraphics[width=0.8\columnwidth]{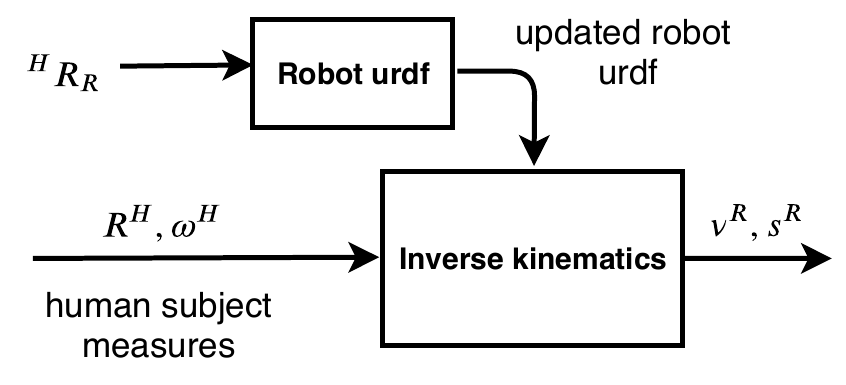}
	\caption{Block diagram of kinematic whole-body motion retargeting.}
	\label{fig:WB-retargeting}
\end{figure}

We perform the whole-body retargeting by geometrically mapping anthropomorphic motions of human links to corresponding robot links. Fig. \ref{fig:WB-retargeting} shows our proposed method for the whole-body retargeting of human motions. Similar to the task space retargeting introduced in Section \ref{sec:kinematic-retargeting}, we formulate the retargeting problem as an inverse kinematics problem given only the rotation and angular velocity $R^H, \omega^H $ of the human links and the robot's URDF model. In our case, the customized mapping between each link of the human and the robot is done with a constant rotation $\prescript{\mathcal{H}}{}{R}_\mathcal{R}$, and applied directly on the robot's URDF for the ease of implementation. An additional benefit with this approach is the consideration of the robot's link properties and joint types using the URDF model. Therefore, the change of the human subject or robot geometry does not affect the retargeting of the human motions to the robot, i.e., the proposed retargeting method increases the scalability enabling application to different human subjects or robots with minimal efforts.

In this formalization, the frame in which an individual human link rotation and angular velocity measurements are expressed should coincide with the corresponding robot link's frame. As an example, the actual link frame definitions of both the human links and the robot links are highlighted in the Fig.~\ref{fig:architecture}.
The frame equivalence from the human to the robot links is indicated by the numbering.
In this case, by identifying the relative rotation between the human link frames and the corresponding robot link frames manually, we obtain the robot's \textit{desired} motion in the frame attached to the robot. Given the rotation from the human link frames to the inertial frame, $\prescript{\mathcal{I}}{}{R}_\mathcal{H}$, and the constant rotation from the robot link frames to human link frames, $\prescript{\mathcal{H}}{}{R}_\mathcal{R}$; equation \eqref{eq:human_robot_transformation} provides the rotation from the \emph{desired} robot link frames to the inertial frame,
\begin{equation}
\prescript{\mathcal{I}}{}{R}^{*}_\mathcal{R}  = \prescript{\mathcal{I}}{}{R}_\mathcal{H} \prescript{\mathcal{H}}{}{R}_\mathcal{R}.
\label{eq:human_robot_transformation}
\end{equation}

The fixed rotation $\prescript{\mathcal{H}}{}{R}_\mathcal{R}$ is computed offline by positioning both the robot and the human models in a similar joint configuration as highlighted in Fig.~\ref{fig:architecture}.

Once the motion of robot links is correctly extracted, the robot joint positions is found by formulating the inverse kinematics as an optimization problem \cite{Sciavicco1988, latella2018towards}. To solve the inverse kinematics problem, we benefit from a dynamical optimization method that ensures the convergence of the frame orientation errors to a minimum \cite{Rapetti2019}. We define the following dynamical system:

\begin{equation}
\mathcal{V} + K \mathcal{E} = 0,
\label{eq:dynamics_Ik}
\end{equation}
in which $K$ is the gain matrix, $\mathcal{E}$ and $\mathcal{V}$ are respectively the vectors collecting orientation and angular velocity errors defined as follow:

\begin{IEEEeqnarray}{CCC}
	\phantomsection
	\IEEEyessubnumber
	\mathcal{E}_i = sk(\prescript{\mathcal{I}}{}{R}^{*}_{\mathcal{R}_i}\prescript{\mathcal{I}}{}{R}^{}_{\mathcal{R}_i})^{\vee},  \\
	\mathcal{V}_i= \prescript{\mathcal{I}}{}{\omega}^{*}_{\mathcal{R}_i}- {J}^{a}_{\mathcal{R}_i}(q) \nu,
	\IEEEyessubnumber
\end{IEEEeqnarray}

where $\mathcal{E}_i, \mathcal{V}_i \in \mathbb{R}^3$ are the errors computed for the i-th link, and $\mathcal{R}_i$ is the i-th link of the robot. Hence, the joint velocities can be found by solving the following optimization problem:
\begin{equation}
\begin{aligned}
& \nu(t) = 
& &  \underset{\nu}{\argmin} \left\lVert \mathcal{V} + K \mathcal{E} \right\rVert  + \left\lVert \lambda \nu \right\rVert, \\
& \text{s.t.} & & G(s) \nu \leq g(s),
\end{aligned}
\label{eq:optimization_Ik}
\end{equation}

in which $\lambda$ is the regularization term, and with $G(s), g(s)$  a linear inequality constraint is defined. Finally, the robot's desired joint positions $s(t)$ are found by integrating $\nu(t)$. To solve the optimization problem we rely on a quadratic programming (QP) library \cite{Stellato2017}.

Our proposed method allows to retarget human motions to a robot even if their kinematics is not similar, i.e., in case the humanoid's robot limb and the human's corresponding limb has different degrees of freedom. Moreover, it enables the anthropomorphic motion retargeting, as the robot mimics human's links motion.

\section{Retargeting Experiments \& Results}
\label{sec:retargeting-experiments-results}

\begin{figure*}[t]
	\centering
	\begin{subfigure}[b]{0.2\textwidth}
		\includegraphics[trim=0cm 10cm 0cm 10cm, clip=true, scale=0.075]{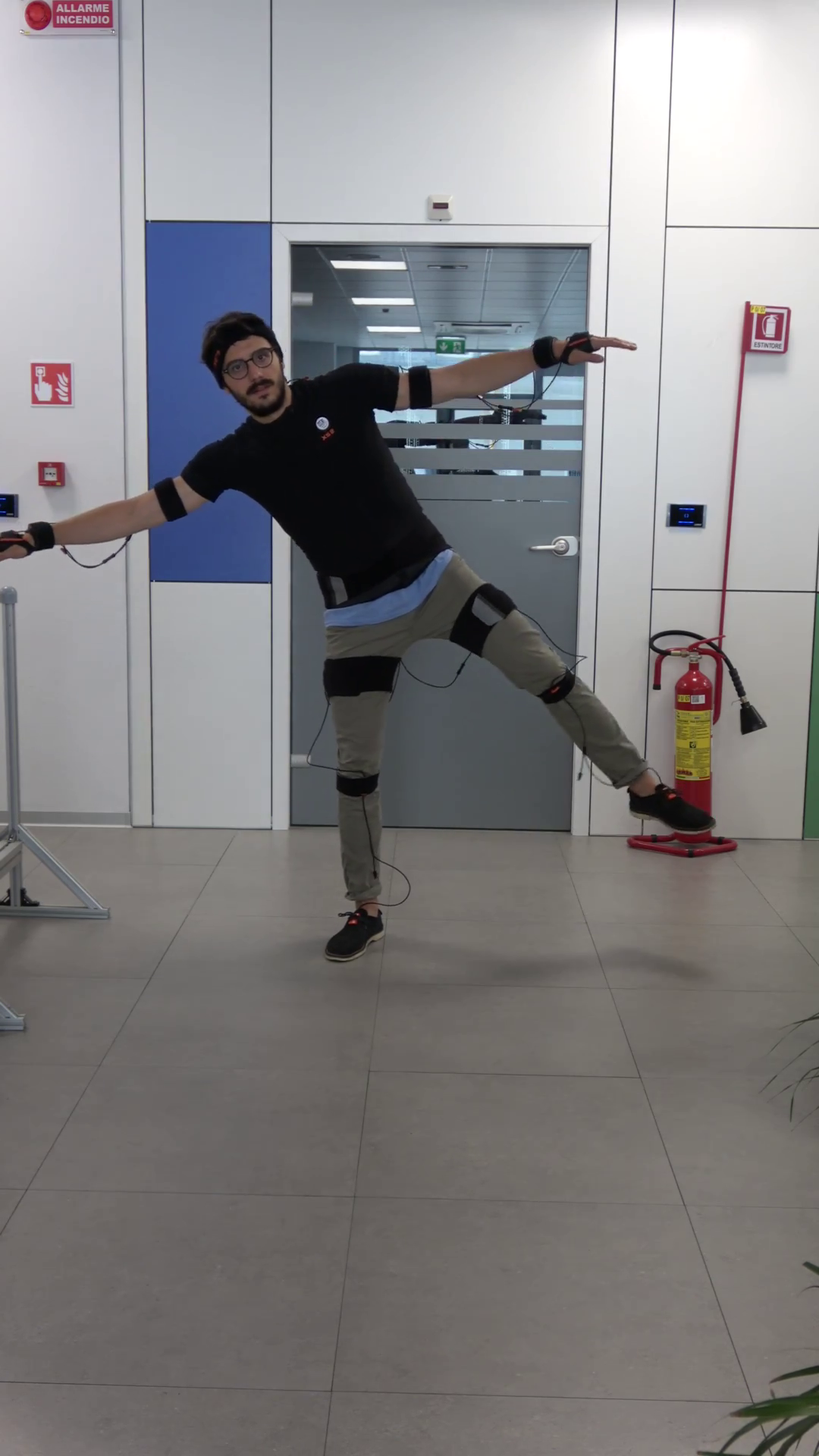}
		\label{fig:rviz-subject-one-foot}
	\end{subfigure}
	~
	\begin{subfigure}[b]{0.775\textwidth}
		\includegraphics[scale=0.1325]{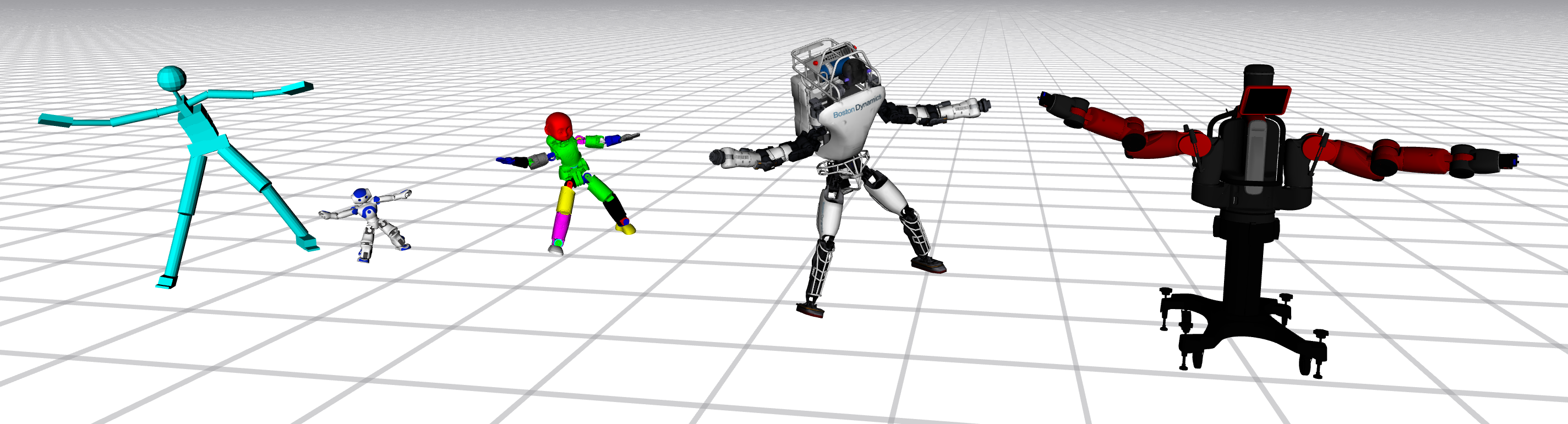}
		\label{fig:rviz-human-one-foot}
	\end{subfigure}
	\begin{subfigure}[b]{0.2\textwidth}
		\includegraphics[trim=0cm 10cm 0cm 10cm, clip=true, scale=0.075]{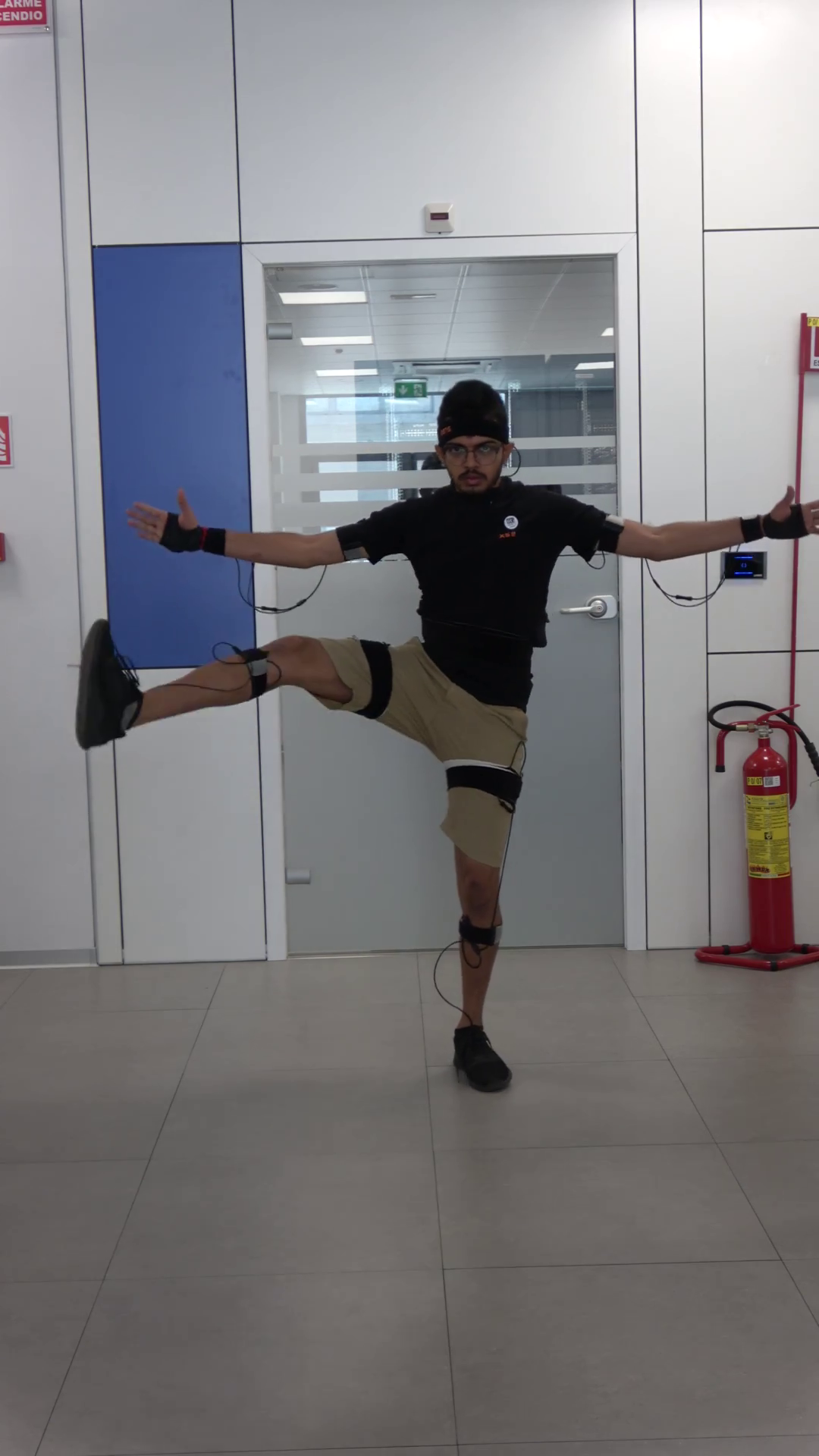}
		\label{fig:rviz-subject-jump}
	\end{subfigure}
	~
	\begin{subfigure}[b]{0.775\textwidth}
		\includegraphics[scale=0.15]{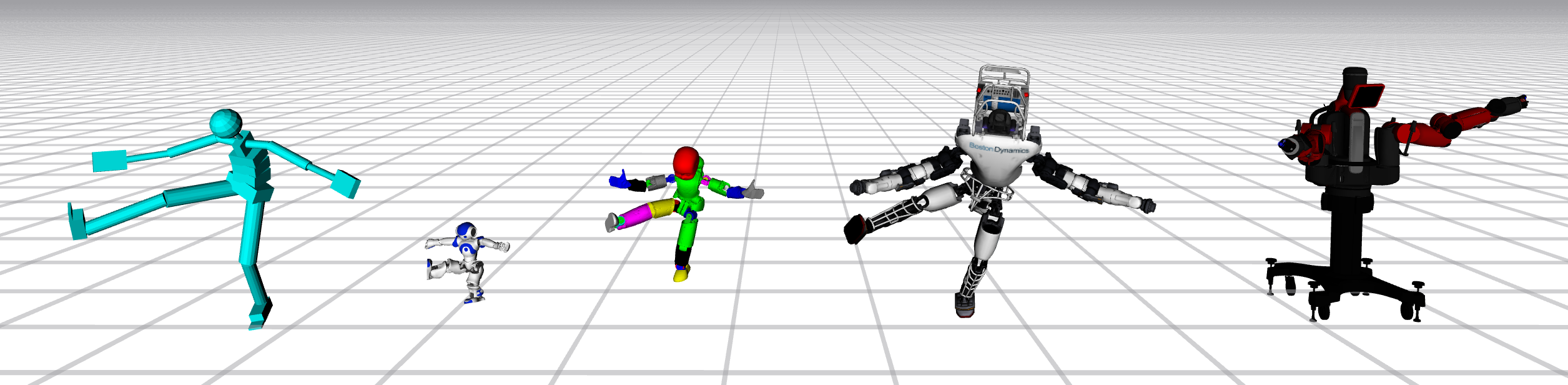}
		\label{fig:rviz-human-jump}
	\end{subfigure}
	\caption{Rviz visualization of whole-body retargeting of  human subjects motion to different models: a) Human Model b) Nao c) iCub d) Atlas e) Baxter; top: human subject stands on the right foot, bottom:  human subject stands on the left foot.}
	\label{fig:scalability}
\end{figure*}

The human limbs' rotation and angular velocity are captured in real time using Xsens motion capture technology that involves several MEMS based inertial sensors placed on various body parts of the human. The whole-body retargeting experiments are performed with motion data captured for two human subjects. To demonstrate the scalability and usability of our proposed method, we perform kinematic retargeting with robots having different degrees of freedom (DoFs). The robot models we considered are a) iCub humanoid robot with 32 DoFs, b) NAO humanoid robot with 24 DoFs, c) Atlas humanoid robot with 30 DoFs. To show that our method is not limited to humanoid robots, we perform a retargeting scenario with Baxter dual arm 15 DoFs robot. Additionally, we show the retargeting with a human model that has 66 DoFs. The Fig.~\ref{fig:scalability} highlights kinematic retargeting with different models and human subjects shown using Rviz kinematic visualization tool. The first row corresponds to the human motion of standing on right foot from the first subject and the second row corresponds to the human motion of standing on the left foot by the second subject. Concerning the baxter robot, the retargeting is done only for the arms and the head, as it is a fixed base robot.

\begin{figure}[t]
	\centering
	\begin{subfigure}[b]{\columnwidth}
		\hspace{0.175cm} \includegraphics[trim=1.6cm 0cm 2.5cm 0cm, clip=true, scale=0.25]{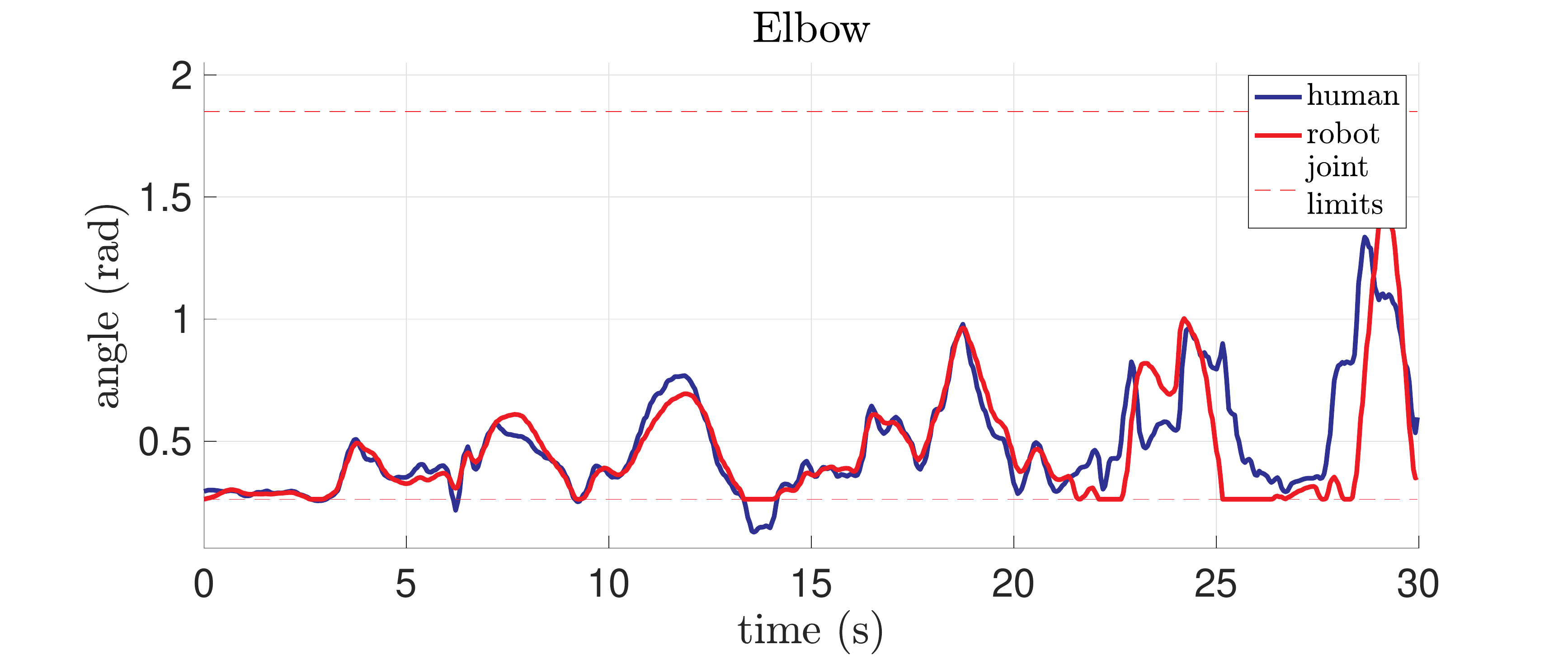}
		\label{fig:error-joint-value} 
	\end{subfigure}
	\begin{subfigure}[b]{\columnwidth}
		\hspace{0.175cm} \includegraphics[trim=1.6cm 0cm 2.5cm 0cm, clip=true, scale=0.25]{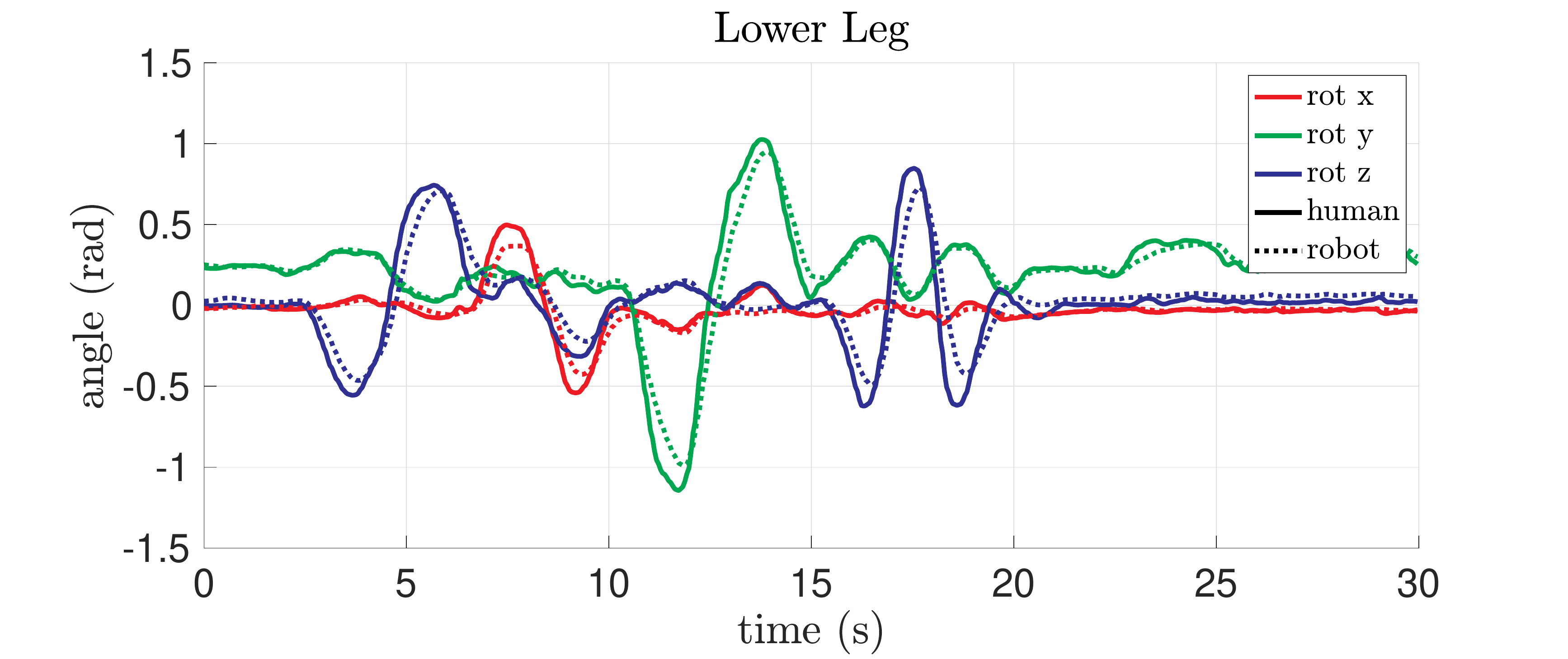}
		\label{fig:error-link-value}
	\end{subfigure}
	\caption{Performance of the whole-body retargeting of the human motions to the iCub robot.
	}
	\label{fig:link_orientation_error_angles_robot_limits}
\end{figure}

The orientation of the human links $\prescript{\mathcal{I}}{}{R}_\mathcal{H}$ is obtained from Xsens measurements. The solution of the inverse kinematics problem formulated in Section \ref{sec:human-motion-retargeting} along with the robot model provides the joint values and velocities of the robot. To compute the robot's achieved link orientation $\prescript{\mathcal{I}}{}{R}_\mathcal{R}$, we use floating base forward kinematics employing the robot's joint values.
Fig.~\ref{fig:link_orientation_error_angles_robot_limits} on the top, shows the human's and the robot's right arm elbow joint values. 
Indeed, when the user moves his elbow in a configuration that it is not feasible for the robot, as can be seen at time instant $t\sim \SI{13}{\second}$, the inverse kinematics finds a feasible solution that tries to minimize the rotation error between the human frames and the iCub robot frames, see \eqref{eq:optimization_Ik}. 
Moreover, according to the robot model, the robot's kinematics may not resemble the human's corresponding ones between two consecutive links, e.g., the robot has lower DoFs than the human, or the order/orientation of joints between the two consecutive links are different. In this case, we use $\prescript{\mathcal{I}}{}{R}_\mathcal{H} \prescript{\mathcal{I}}{}{R}_\mathcal{R} ^\top$ measure to evaluate the error between the robot's link orientation and the human's one. Fig.~\ref{fig:link_orientation_error_angles_robot_limits} on the bottom shows the human's right lower leg link rotation matrix and corresponding one of the iCub robot computed through kinematic whole-body retargeting.
For the sake of comprehension, the rotation matrix is parametrized using \textit{Euler angles} expressed a series of $x$,$y$,$z$ intrinsic rotations.  

\section{Teleoperation Experiments \& Results}
\label{sec:teleoperation-experiments-results}

Towards demonstrating the capabilities of our whole-body retargeting, we perform teleoperation experiments using two state-of-the-art \textit{whole-body controllers} for humanoid robots. The whole-body teleoperation experiments are carried with the 53 degrees of freedom iCub robot that is $\SI{104}{\centi \meter}$ tall \cite{natale2017icub}.
The controllers run at $\SI{100}{\hertz}$ while the retargeting application runs at $\SI{200}{\hertz}$.
The average walking speed of the robot is $\SI{0.23}{\meter \per \second}$.
Both the applications are run on a machine of $4$th generation Intel Core i7@$\SI{1.7}{\giga \hertz}$ with 8GB of RAM.

\subsection{Whole-Body Teleoperation with Balancing Controller}
\label{sec:balancing-controller}

Momentum-based control \cite{nava2016stability} \cite {herzog2014balancing} proved to be effective for maintaining the robot's stability by controlling the robot's momentum as the primary objective. Additionally, a \textit{postural task} projected into the nullspace of the primary task can be used for performing additional tasks like manipulation while ensuring the stability of the robot. The control problem is formulated as an optimization problem to achieve the two tasks while carefully monitoring and regulating the contact wrenches, considering the associated feasible domains by resorting to quadratic programming (QP) solvers.

\begin{figure*}[!ht]
	\centering
	\begin{subfigure}[b]{0.235\textwidth}
		\includegraphics[width=\textwidth]{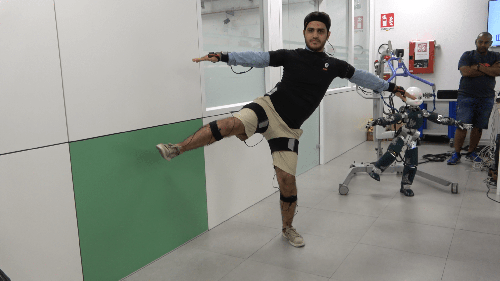}
		\caption{}
		\label{fig:balancing1} 
	\end{subfigure}
	~
	\begin{subfigure}[b]{0.235\textwidth}
		\includegraphics[width=\textwidth]{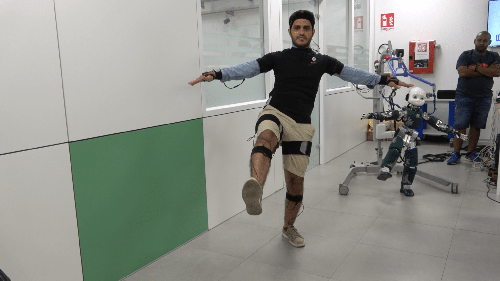}
		\caption{}
		\label{fig:balancing2}
	\end{subfigure}
	~
	\begin{subfigure}[b]{0.235\textwidth}
		\includegraphics[width=\textwidth]{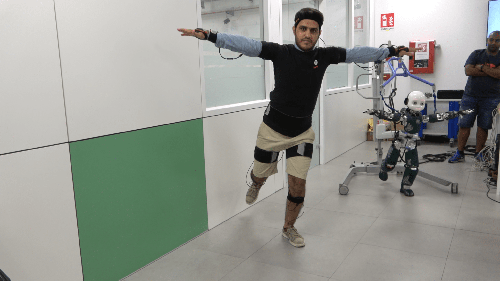}
		\caption{}
		\label{fig:balancing3}
	\end{subfigure}
	~
	\begin{subfigure}[b]{0.235\textwidth}
		\includegraphics[width=\textwidth]{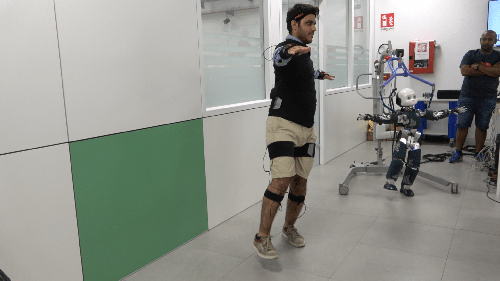}
		\caption{}
		\label{fig:balancing4}
	\end{subfigure}
	\caption{Whole-body retargeting with balancing controller snapshots}
	\label{fig:balancing-controller-snapshots}
\end{figure*}

We considered one such momentum-based balancing controller \cite{nava2016stability} and extended the postural task by giving the joint references from whole-body retargeting.
Fig.~\ref{fig:balancing-controller-snapshots} shows snapshots from the experiments of the whole-body retargeting with the balancing controller.  

\begin{figure*}[!ht]
	\centering
	\begin{subfigure}[b]{\textwidth}
		\centering
		\hspace{-0.225cm} \includegraphics[height=10em,width=43.25em]{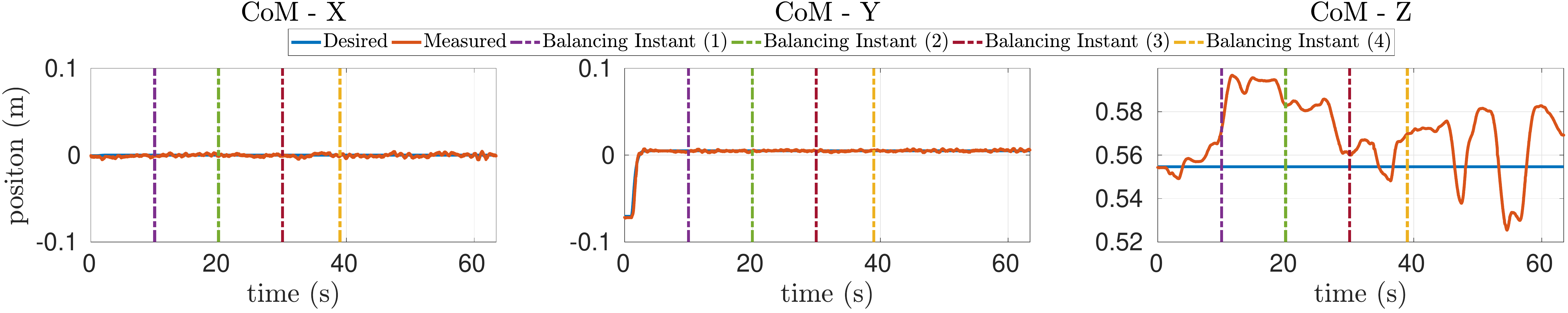}
		\label{fig:balancing_com_tracking}
	\end{subfigure}
	\vskip 0.1cm 
	\begin{subfigure}[b]{\textwidth}
		\centering
		\includegraphics[height=8.5em]{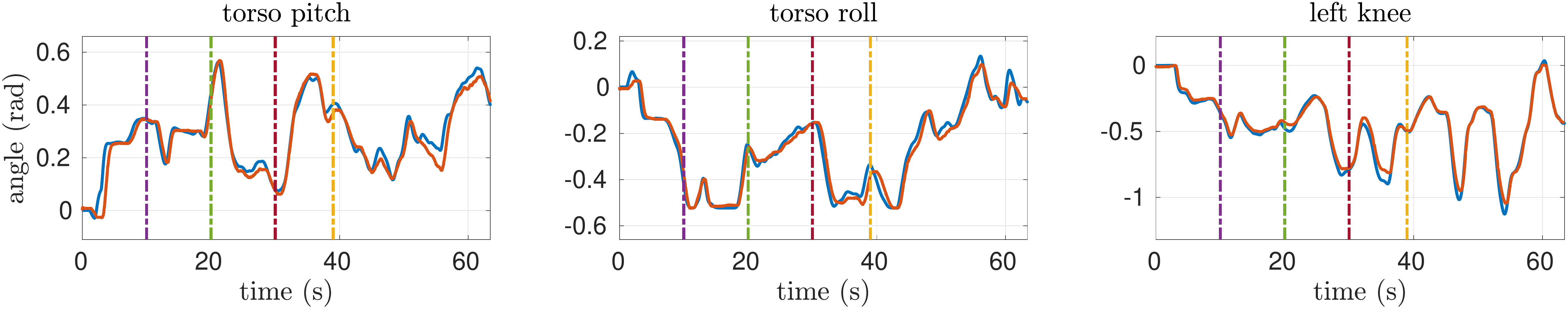}
		\label{fig:balancing_torso_tracking}
	\end{subfigure}
	\vskip 0.1cm 
	\begin{subfigure}[b]{\textwidth}
		\centering
		\includegraphics[height=8.5em]{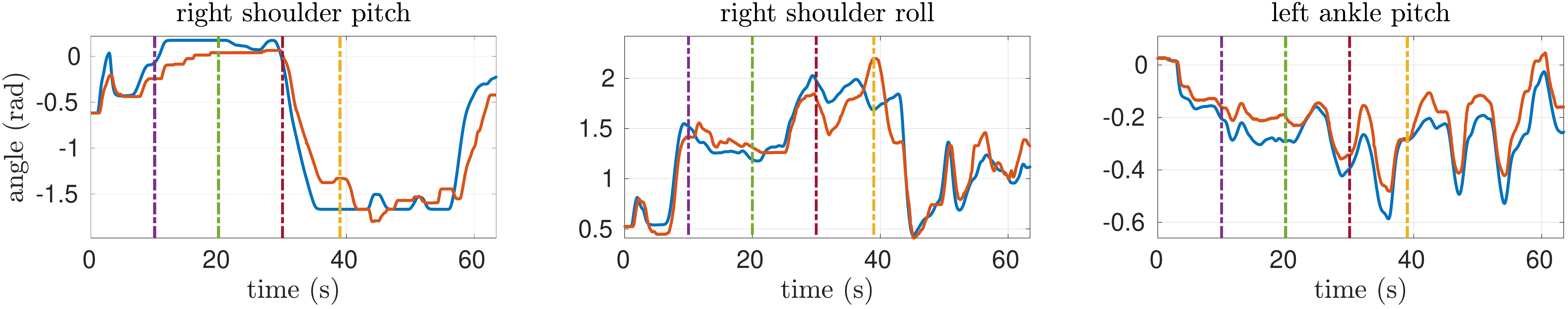}
		\label{fig:balancing_right_arm_tracking}
	\end{subfigure}
	\caption{Tracking of the center of mass and the joint angles during the whole-body retargeting with balancing controller; blue line represents the desired quantity, orange line is the actual robot quantity.}
	\label{fig:balancing_joint_tracking}
\end{figure*}

In this experiment the robot is balancing on the left foot and maintaining the stability of its center of mass as shown in Fig.~\ref{fig:balancing_joint_tracking}. Additionally, it tracks all the joints with the references coming from whole-body retargeting.
The vertical dashed lines correspond to the experimental snapshots indicated in Fig.~\ref{fig:balancing-controller-snapshots}.
The references to the $x$ and $y$ components of the CoM are close to zero to maintain the stability of the robot by keeping the CoM inside the support polygon and the gains are tuned to achieve good tracking. The CoM motion along the $z$-axis does not effect the stability of the robot and the gain value of the $z$ components is kept lower in order to allow the vertical movements of the robot during retargeting. The input joint references from retargeting are smoothed through a minimum-jerk trajectory  \cite{Pattacini2010a}.
A smoothing time parameter is tuned in order to achieve good balancing between the postural tracking and stability. Accordingly, the joints such as torso pitch, torso roll, and left knee for which the human does not move fast, while balancing on left foot, achieve good tracking. On the other hand, the joints such as right shoulder pitch, right shoulder roll, and left ankle pitch are moved frequently while performing the retargeting and hence the tracking is not close owing to the delay from the smoothing time involved in producing minimum-jerk trajectory joint references for the robot joints. Ideally, the smoothing time can be kept lower considering that we receive continuous joint references from retargeting. At this point, we did not conduct exhaustive tests to find the lower threshold for the smoothing parameter that ensures fast and accurate retargeting of dynamic motions from the human while maintaining the robot's stability.

\subsection{Whole-Body Teleoperation with Walking Controller}
\label{sec:walking-controller}

Humanoid robot walking is another challenging control paradigm. Divergent-Component-of-Motion (DCM) based control architectures proved promising for humanoid robot locomotion \cite{Romualdi2018, Englsberger2015}. The architecture typically consists of three layers: 1) Trajectory generation and optimization layer that generates the desired footsteps and the DCM trajectories \cite{Englsberger2015}; 2) Simplified model control layer that implements an \emph{instantaneous} control law with the objective of stabilizing the unstable DCM dynamics; and 3) Whole-body control layer that guarantees the tracking of the robot's set of tasks, including the Cartesian tasks and the postural tasks, using the stack-of-tasks paradigm implemented through a quadratic programming (QP) formalism.

\begin{figure*}[!t]
	\centering
	\begin{subfigure}[b]{0.3\textwidth}
		\includegraphics[width=\textwidth]{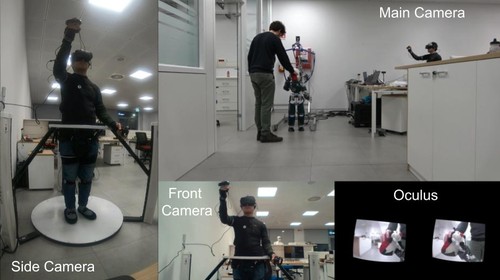}
		\caption{first stage}
		\label{fig:first-stage} 
	\end{subfigure}
	~
	\begin{subfigure}[b]{0.3\textwidth}
		\includegraphics[width=\textwidth]{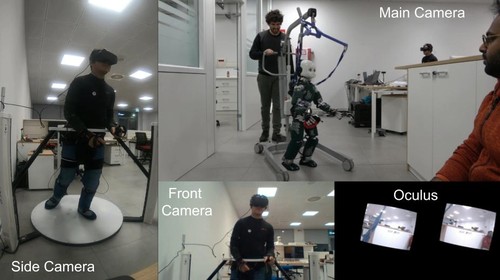}
		\caption{second stage}
		\label{fig:second-stage}
	\end{subfigure}
	~
	\begin{subfigure}[b]{0.3\textwidth}
		\includegraphics[width=\textwidth]{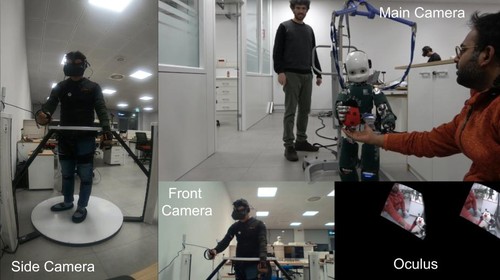}
		\caption{third stage}
		\label{fig:third-stage}
	\end{subfigure}
	\caption{Whole-body retargeting with walking controller experimental stages}
	\label{fig:walking-controller-experimental-stages}
\end{figure*}

We considered one such DCM based walking controller \cite{Romualdi2018} and extended the postural task by giving the joint references from whole-body retargeting. Fig.~\ref{fig:walking-controller-experimental-stages} shows the three different experimental stages of whole-body retargeting with the walking controller. During the first and the third stages the robot is in double support standstill phase while during the second stage the robot is in walking phase.

\begin{figure*}[!t]
	\centering
	\begin{subfigure}[b]{\textwidth}
		\centering
		\hspace{0.1cm} \includegraphics[height=10em, width=42.5em]{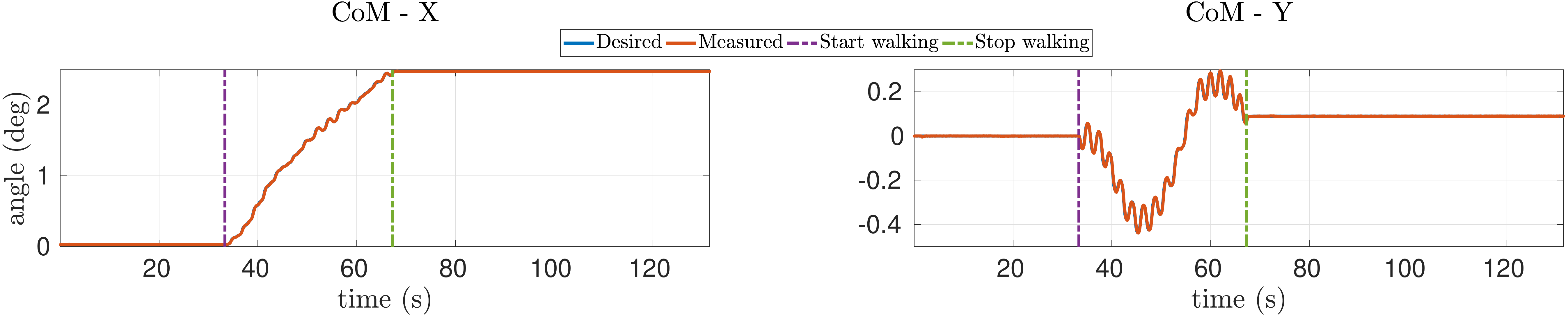}
		\label{fig:walking_com_tracking}
	\end{subfigure}
	\vskip 0.1cm
	\begin{subfigure}[b]{\textwidth}
		\centering
		\includegraphics[height=8.5em]{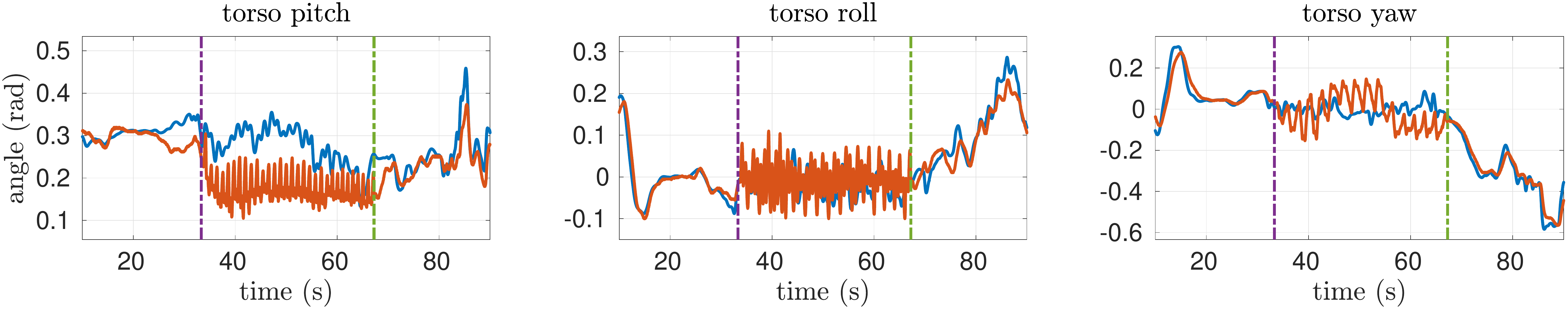}
		\label{fig:walking_torso_tracking}
	\end{subfigure}
	\vskip 0.1cm
	\begin{subfigure}[b]{\textwidth}
		\centering
		\includegraphics[height=8.5em]{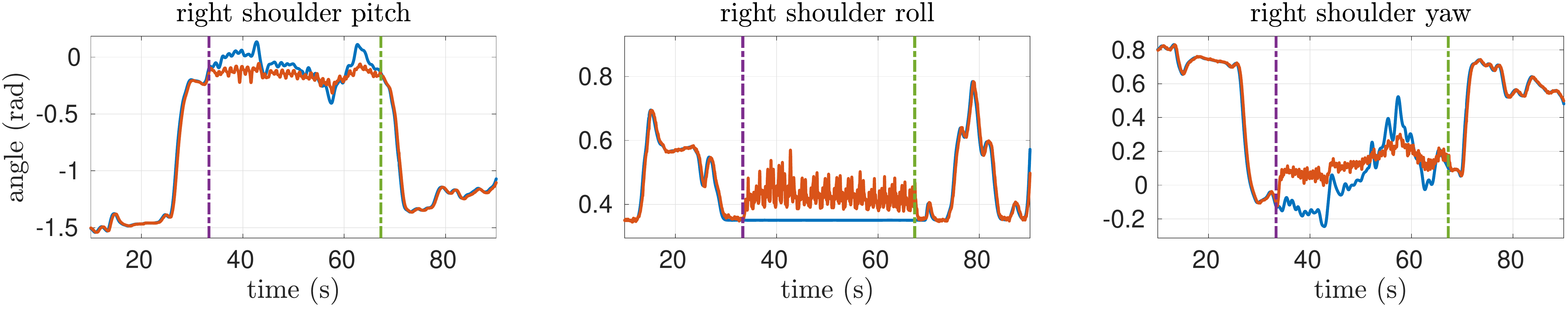}
		\label{fig:walking_right_shoulder_tracking}
	\end{subfigure}
	\caption{Tracking of the center of mass and the joint angles during the whole-body retargeting with walking controller.}
	\label{fig:walking_joint_tracking}
\end{figure*}

The walking controller's primary objective is to track the center of mass $x$ and $y$ components along the desired trajectory. The overall center of mass tracking of the $x$ and $y$ components is very good for the entire duration of the experiment as shown in Fig.~\ref{fig:walking_joint_tracking}.

Currently, we engage only the upper body retargeting, and the lower body is controlled by the walking controller. During our experiments we observed that the weights for achieving satisfactory upper body retargeting of the postural task during the double support standstill phase and the walking phase are different. Having the same retargeting gains while walking lead to uncoordinated movements eventually compromising the robot's stability while walking. So, we choose higher retargeting gains during double support standstill phase and the gain values are set to zero during the walking phase. The transition between the two sets of weights is achieved smoothly through minimum jerk trajectories \cite{Pattacini2010a}. Fig.~\ref{fig:walking_joint_tracking} highlights tracking for some of the upper-body joints. The blue line represents the desired joint position provided by human motion retargeting and the orange line is the actual robot joint position. The purple vertical dashed line indicates the starting instance of the \emph{second stage}, i.e., walking, and the green vertical dashed line indicates the stopping instance of walking phase. During the \emph{first stage}, human motion retargeting is good and the joint position error is low. Instead, during the \emph{second stage}, as the robot starts walking the joint position error is higher as the retargeting gains are set to zero.

\section{Conclusions \& Future Work}
\label{sec:conclusions}

In this paper, we propose and validate a whole-body teleoperation framework for humanoid robots,  leveraging the geometric retargeting of motion from human body parts to the analogous humanoid robot parts. The proposed approach increases the usability by employing solely the robot's model and by considering the orientation and angular velocity measurements from the human links.

The proposed retargeting approach has been applied to multiple robot models using motion data from multiple human subjects.
Furthermore, we performed active retargeting experiments during bipedal balancing and locomotion tasks using two state-of-the-art whole-body controllers for humanoid robots. Our experimental validation strongly supports our proposed framework.

Currently, in the balancing controller the center of mass references are independent from the human while the retargeting is done in the postural space. 
Additionally, in the walking controller, we restrict ourselves to do only upper-body retargeting in postural space. The center of mass and feet trajectory references are independent from the human.
In the future work, we will extend our framework to address the above limitations.

\bibliography{bibliography}
\end{document}